%% file: main.tex
\documentclass[letterpaper]{article}

\usepackage{amsmath}
\usepackage{hyperref}
\usepackage{cleveref}
\usepackage{natbib,alifeconf}  
\usepackage{url}
\usepackage{booktabs}
\usepackage[export]{adjustbox}
\usepackage{subcaption}

\usepackage[utf8]{inputenc}
\usepackage{xcolor}
\usepackage{graphicx} 

\title{Time to Play: Simulating Early-Life Animal Dynamics\\
Enhances Robotics Locomotion Discovery }

\author{
    Paul Templier$^{1}$,
    Hannah Janmohamed$^{1}$,
    David Labonte$^{2}$, 
    Antoine Cully$^{1}$ \\
    \mbox{}\\
    $^1$Adaptive \& Intelligent Robotics Lab, Imperial College London, United Kingdom \\
    $^2$Evolutionary Biomechanics Laboratory, Imperial College London, United Kingdom \\
    p.templier@imperial.ac.uk
} 

\begin{document}

\maketitle

\begin{abstract}

Developmental changes in body morphology profoundly shape locomotion in animals, yet artificial agents and robots are typically trained under static physical parameters. 
Inspired by ontogenetic scaling of muscle power in biology, we propose \emph{Scaling Mechanical Output over Lifetime (SMOL)}, a novel curriculum that dynamically modulates robot actuator strength to mimic natural variations in power-to-weight ratio during growth and ageing. 
Integrating SMOL into the MAP-Elites quality-diversity framework, we vary the torque in standard robotics tasks to mimic the evolution of strength in animals as they grow up and as their body changes.
Through comprehensive empirical evaluation, we show that the SMOL schedule consistently elevates both performance and diversity of locomotion behaviours across varied control scenarios, which we hypothesise to be thanks to agents leveraging advantageous physics early on to discover skills that act as stepping stones when they reach their final standard body properties. Based on studies of the total power output in humans, we also implement the SMOL-Human schedule that models isometric body variations due to non-linear changes like puberty, and study its impact on robotics locomotion.


\end{abstract}


Code available at: \url{https://github.com/TemplierPaul/SMOL}

\begin{figure}[h]
    \centering
    \includegraphics[width=0.9\linewidth]{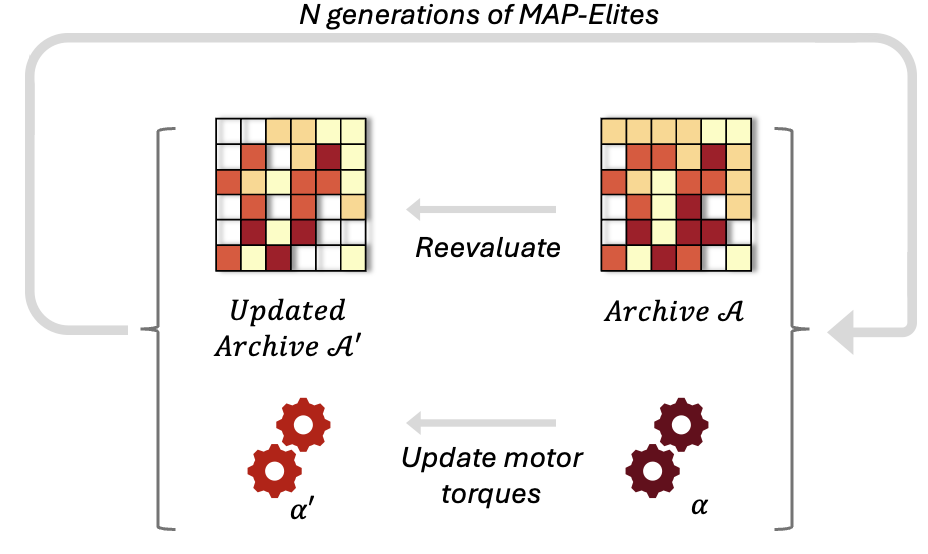}
    \caption{Overview of our schedule-based MAP-Elites framework. After each fixed interval of generations, the actuator torque scaling factor $\alpha$ is updated to a new value $\alpha{\prime}$ according to a predefined schedule. The current archive $\mathcal{A}$ is then fully reevaluated under the new physical model and transferred into a new archive $\mathcal{A}{\prime}$, which seeds the next evolutionary batch. This process allows behaviour adaptation to changing embodiment over time and enables the study of developmental trajectories in quality-diversity search.
}
    \label{fig:teaser}
\end{figure}

\section{Introduction}

Human and animal locomotion can change significantly across a lifespan as the body and brain mature.
A child, a young adult, and an elderly person may each walk, run, and balance differently—not only because of changes in coordination or control, but also due to changes in physical capacity, such as the force that their muscles can generate relative to their weight.
Across a lifespan, this power-to-weight ratio undergoes significant transformations, but humans do not need to relearn to walk from scratch every time. Instead, humans and animals learn to move with a changing body, using previously discovered locomotion methods to adapt to new constraints. The initial learning phase happens during childhood, when younglings can be seen exploring their environment by running and jumping around to play, hence developing diverse locomotion gaits \citep{infant_locomotion}.

\subsection{Time to play}
Play behaviour has been observed in many species from mammals to birds and insects, and has been theorized to be critical for the development of animal instincts and skills \citep{groosplay,play_evolution}. Playing at a young age helps learn diverse locomotion skills, each serving as a stepping stone to develop new walking patterns \citep{infant_locomotion}.
Some species like Assamese macaques invest energy into play rather than prioritizing physical growth, showing the importance of early-life locomotor development \citep{play_growth_tradeoff}. 


Play behaviour is predominantly observed in young animals \citep{hill2014adults,shimada2014importance}, disappearing when they reach adulthood. This skill learning phase is accompanied by physical changes in body size and properties \cite{channon2019ontogenetic} that influence their muscular strength, and hence their behavioural envelope as the set of actions that are physically possible in the environment.



By contrast, robots and artificial agents are typically trained and evaluated under similar physical parameters; no allowance is made for physical development or decline. 
Most robots are designed with fixed mechanical properties, such as constant motor torque, and most learning algorithms assume the agent’s body stays the same over time.
While methods like domain randomisation \citep{tobin2017domain, chen2021understanding} vary physical properties to improve the final adaptability of the robot to the real world, they still consider the robot to be in a globally stable configuration.


\subsection{Scaling Mechanical Output over Lifetime}

In this work, we take inspiration from the observed correlation between skill learning and ontogenetic scaling to explore how changing physical properties during training influences the quality and diversity of locomotion behaviours discovered by evolutionary robotics methods. 
To this end, we introduce the \textbf{Scaling Mechanical Output over Lifetime} (\textbf{SMOL)} approach, which simulates changes in strength by modulating a robot’s motor torque during training to loosely mimic the natural progression of muscular capacity across animal lifespans. 
Using SMOL, we investigate how developmentally-inspired physical variation, coupled with continuous behavioural adaptation, can shape the emergence of diverse and effective walking gaits.

To support adaptation across these physical variations, we use quality-diversity (QD) algorithms \citep{qdunifying, book_chapter} to learn control policies for our robots in simulations.
QD methods are evolutionary algorithms that search for a wide variety of effective behaviours, and maintain a repertoire of them called \textit{archive}.
This archive plays a crucial role in our approach: when the robot’s strength profile changes, learning does not restart.
Instead, evolution continues from previously discovered behaviours in the archive.
In this way, the archive functions as a developmental memory, similar to how animals may adapt their movements over time based on prior experience.

By combining developmental torque evolution with continuous quality-diversity search, we investigate how time-varying physical capabilities influence the quality and diversity of the learned walking gaits. We define nature-inspired schedules to scale the actuator torque in robots from three standard control environments, effectively changing their power-to-weight ratio while they learn to develop locomotion skills, and find they outperform the standard approach of constant actuator strength.


When tested on the same final environment, schedules that mimic biological patterns, such as a gradual decline resembling human ageing or animal senescence, lead to significantly better performance in both the quality and diversity of learned behaviours than standard approaches that keep the model constant, across all tasks.

\section{Background and Related Works}
\subsection{Muscle Scaling Laws and Human Ageing}
Muscle strength in vertebrates and invertebrates originates from muscle contraction \citep{flight_muscle}.
The total force a muscle can generate scales directly with the number of fibres, and hence its cross-sectional area \citep{mcmahon1984muscles}.
Animal mass, however, depends on body volume.
If we denote a characteristic linear size by $s$, the cross-sectional area grows as $s^2$ and the volume as $s^3$. 
Hence smaller animals possess a higher force‐to‐mass ratio $s^2/s^3 = s^{-1}$\citep{schmidt1984scaling,lindstedt1987allometry}. 
Thus, as animals grow, their theoretical strength per kilogram should diminish, reducing their accessible musculoskeletal performance space \citep{labonte2023theory}.

In practice, muscle performance may not follow this theoretical scaling due to influences such as hormonal changes, connective tissue stiffness \citep{holt2016stuck}, changes to the muscle mass to body mass ratio, and allometric changes in the development and growth of limbs and organs \citep{channon2019ontogenetic}. In humans, the total mechanical output per unit mass does not vary monotonically: it rises sharply during early development, peaks in early adulthood, and subsequently declines with age \citep{power_lifespan}. 

In this paper, we aim to loosely mimic these effects in robotic agents by varying joint torque responses to control signals over time, to simulate changing power-to-weight ratios; this is described in detail in the Methodology section. 

\subsection{Quality-Diversity}

Traditional optimisation algorithms typically aim to find a single high-performing solution to a given task.
In contrast, Quality-Diversity (QD) algorithms seek to discover a wide range of effective solutions that differ meaningfully, not just in how well they perform, but in how exactly the performance is achieved (behaviour) \citep{qdunifying, book_chapter}.
In these algorithms, solutions are not only assessed via a fitness function, but are also characterised by their features (also referred to as \textit{measures} or \textit{behaviour descriptors} in literature \citep{book_chapter, qdunifying}).
The goal of QD algorithms is to find solutions which are both high-performing and also diverse according to these features.
Such a more nuanced approach is particularly valuable in domains like robotics, where diverse strategies can improve robustness, adaptability, and long-term learning potential \citep{nature, hbr}.

One of the most widely used QD algorithms is MAP-Elites \citep{mapelites}.
MAP-Elites maintains a structured archive, denoted $\mathcal{A}$, which discretises this behavioural space into a grid of cells or niches.
Each cell in $\mathcal{A}$ stores the as-of-now best-performing solution that falls within its descriptor bounds.
As the evolutionary process generates new candidates, MAP-Elites evaluates both their task performance and behavioural descriptor.
If a candidate either outperforms the current elite in its niche or fills a previously unoccupied niche, it is added to $\mathcal{A}$.

This approach encourages the discovery of both high-quality and diverse solutions across the behaviour space.
One of its key strengths is its ability to preserve stepping-stone solutions—intermediate behaviours that may not be globally optimal, but are crucial for reaching better solutions in future iterations \citep{qdsteppingstones, roleofsteppingstones}.
Since the archive $\mathcal{A}$ retains a variety of such behaviours, it enables evolutionary processes to revisit and build upon them later.

In our work, we leverage this property to support learning in developmentally dynamic robots.
As the robot’s power-to-weight ratio changes over time through torque schedules, the archive $\mathcal{A}$ serves as a memory of previously successful behaviours.
Rather than restarting learning each time the body changes, evolution continues from these stored behaviours, adapting them to new physical constraints and allowing for continuity across developmental stages.

\subsection{Changing the environment}
In robotics, domain randomisation introduces random variability in simulation parameters such as mass or friction, to improve policy robustness and transfer to real-world settings \citep{tobin2017domain, peng2018sim, sadeghi2016cad2rl}.
This has been domain widely used in sim-to-real learning, often without explicit structure.
More recent works explore adaptive or curriculum-based randomisation to better support learning \citep{mehta2020active, openai2019rubik}.


Curriculum learning studies the impact of the order in which samples are seen by the optimization method on the performance \citep{soviany2022curriculum}. It has been successfully applied to Reinforcement Learning, to select the order in which agents will encounter a set of tasks \citep{narvekar2020curriculum}. This often involves setting intermediate sub-goals of shaping the reward function to build increasingly harder tasks that help solve the hardest one, focusing on the problem at hand rather than on the robot's properties \citep{portelas2020automatic}.

By contrast, several works  in evolutionary robotics and artificial life do explicitly consider jointly evolving morphology and control
\citep{lipson2000automatic, kriegman2018morphological, schaff2019jointly}. These works show that optimising the body and controller together can lead to more capable agents, but highlight the fragility of learned behaviours body changes.
Some approaches introduce gradual morphological change or developmental transitions to improve continuity and transfer \citep{auerbach2014environmental, nadizar2022schedule, vujovic2017evolutionary, kriegman2018morphological}.
Our approach shares similarities with body-brain co-evolution methods, which explore how control adapts to changing morphologies; however, unlike those works that aim to discover optimal bodies for a task, we fix the target body and use scheduled changes in torque as a developmental scaffold, ultimately optimising policies for a standard, shared morphology.

\section{Methodology}

\subsection{General Framework}
We study how variations in force-to-mass ratio influence the diversity and performance of behaviours learned through evolutionary methods.
To do this, we embed time-varying actuator torque schedules into a MAP-Elites framework applied to continuous control tasks involving legged locomotion (\Cref{fig:teaser}).

We introduce this as the \textbf{Scaling Mechanical Output over Lifetime} (\textbf{SMOL)} approach, which simulates developmental or physiological change during an agent’s training lifetime.
The following sections describe how we implement and apply these schedules, and how they are integrated into the MAP-Elites process.

\subsection{Power-Weight Ratio Schedules}
\label{sec:power-weight-simulation}

\begin{figure}
    \centering
    \includegraphics[width=0.8\linewidth]{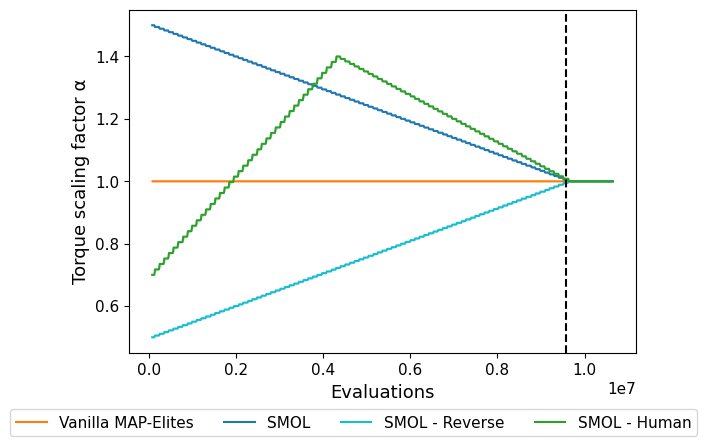}
    \caption{Illustration of the torque scaling schedules used to simulate changes in power-to-weight ratio over time. Each curve represents a different developmental trajectory applied to the robot’s actuators by adjusting the torque scaling factor $\alpha$. The standard schedule (Vanilla MAP-Elites) keeps torque constant throughout training. The SMOL and SMOL-Reverse schedules simulate monotonic decline and growth in strength, respectively. The SMOL-Human schedule follows a biologically inspired pattern: increasing from childhood, peaking in early adulthood, and gradually declining—reflecting empirical human muscle power trends. All schedules are set to $\alpha = 1$ in the final phase of training to allow fair comparison under identical physical conditions.
}
    \label{fig:schedules}
\end{figure}

To study how varying power-to-weight ratios affect the behavioural capabilities of robotic agents, we modify the actuation strength of simulated robots in the Brax physics environment \citep{brax}, where agents are controlled through torques applied by their actuators.
Specifically, we multiply the gear ratio parameter\footnote{\href{https://mujoco.readthedocs.io/en/stable/XMLreference.html\#actuator-general-gear}{https://mujoco.readthedocs.io/en/stable/XMLreference.html}} of all actuators by a scaling factor $\alpha$, where $\alpha > 1$ increases the net torque output for a given control input. 
MuJoCo actuators (used in Brax tasks) are defined as \textbf{ideal torque sources}, with no drop-off in torque as speed goes up, which implies there is no power cap on the actuators like there would be on a real muscle or motor. Changing the gear ratio hence only impacts the torque output, and effectively makes the actuator more powerful. This simulates stronger “muscles” without altering the robot’s mass.
These adjustments are made every few generations during the training process according to predefined schedules, which mimic different developmental or physiological patterns.

To ensure a fair comparison across schedules, we fix $\alpha = 1$ during the final 10 phases of evolution.
This standardisation brings all agents to the same physical state, regardless of how their torque capacity changed earlier, and hence provides a consistent basis for evaluation. Conceptually, this could be seen as representing a shared “adulthood” or “mature body” stage, allowing us to assess how different developmental trajectories influence final performance and behavioural diversity under identical conditions.

\Cref{fig:schedules} provides an overview of the schedules tested in this work.
Each schedule defines how $\alpha$ evolves over time, affecting the robot’s torque capacity during training.
We investigate two primary types of schedules:

\subsubsection{Schedule Type 1: Muscle-Scaling Physics Schedule}
\label{sec:muscle-scaling}

This schedule is based on theoretical scaling laws from biomechanics: smaller organisms have a higher power-to-weight ratio due to their higher surface-to-volume ratio.
To simulate this in robots, we vary the torque limits over time:
\begin{itemize}
    \itemsep0em
    \item \textbf{SMOL} starts with a high torque value (here $\alpha = 1.5$), simulating a small, powerful “child-like” agent, and gradually decays to normal torque ($\alpha = 1$) over training.
    \item \textbf{SMOL-Reverse} implements the opposite approach: it starts with a low torque value (here $\alpha = 0.5$), simulating a physically weaker “immature” agent, and increases to normal torque ($\alpha = 1.0$) over time.
\end{itemize}

These changes are implemented as a smooth schedule (i.e. linear here) across 90\% of the full training horizon, with the last 10\% of training using the standard schedule of $\alpha = 1$ so that the final performance is evaluated under equal torque settings to ensure comparability.


\subsubsection{Schedule Type 2: Human Lifespan Power Schedule}
\label{sec:human-schedule}

This schedule is inspired by empirical studies of human mechanical output per kilogram across the lifespan, which follow a characteristic trajectory: increasing through childhood, peaking in early adulthood, and gradually declining with age \citep{power_lifespan}. We simulate this trend using the \textbf{SMOL-Human} torque schedule. The robot begins in a \emph{childhood} phase with $70\%$ of standard torque capacity ($\alpha = 0.7$), then ramps up during \emph{adolescence and early adulthood} to a peak of $140\%$ ($\alpha = 1.4$), reflecting maximal physical strength. Finally, in the \emph{adulthood} phase, torque gradually decreases to a stabilized level of $100\%$ ($\alpha = 1.0$) to mimic adult human muscle capacity.
The curve thus follows a bell-shaped trajectory, aligning with biological data on human muscle performance over time.




\subsection{Continual Development via MAP-Elites}

To simulate the effects of changing physical capabilities over a robot’s lifetime, we integrate torque schedules (described in the previous section) directly into the MAP-Elites evolutionary process. Each schedule defines how the robot’s power-to-weight ratio evolves during training, similar to muscular development and decline in biological organisms.

We split the total number of generations of MAP-Elites into 100 equal phases.
At the end of every phase, the physics model is updated according to the schedule by setting the value of $\alpha$ to a new value $\alpha{\prime}$.
This update simulates a developmental change in the robot’s body—for example, a child becoming stronger, or an adult gradually losing strength.
However, just as humans must adapt their movements when their physical capacity changes, robotic behaviours also change when the underlying body model is altered.
To account for this, we reevaluate all solutions in the current archive $\mathcal{A}$ using the updated value $\alpha{\prime}$.
This step ensures that the archive reflects how each behaviour performs under the new physical conditions. Using their fitness and descriptor values from the new environment, the reevaluated solutions are then stored in a new archive $\mathcal{A}{\prime}$, which becomes the starting point for the next batch of generations, with the new $\alpha{\prime}$.

Archive reevaluation can affect both the fitness and behavioural descriptors of stored solutions.
In some cases, behaviours that previously occupied distinct cells in the archive may  still behave similarly and fall into the same cell.
Since MAP-Elites retains only the highest-performing solution per niche, this can temporarily reduce diversity—but it ensures the archive remains valid under the current embodiment.

\section{Experimental setup}
\label{sec:experiments}

\begin{figure}
    \centering
    \includegraphics[width=0.6\linewidth]{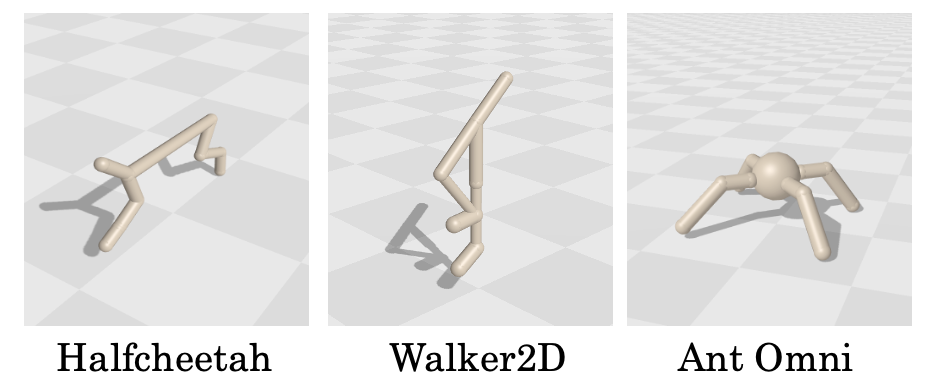}
    \caption{The three Brax locomotion environments used in our experiments: Halfcheetah, Walker2D, and Ant Omni.}
    \label{fig:tasks}
\end{figure}


We evaluate our approach using three continuous control tasks in the Brax v2 physics simulator \citep{brax}, leveraging the distributed QDax framework \citep{qdax} to facilitate parallelisation of MAP-Elites.
To support dynamic torque schedules, we use Brax v2 with the spring physics backend, which offers improved speed and reduced sensitivity to GPU-related noise compared to other backends.\footnote{This may explain differences in scores reported in other works that use a different Brax version, backend, or setup, or use MuJoCo or Gym environments. All experiments presented here are however run using the exact same physics and implementations for consistency.} 
Each robot is controlled by a feedforward neural network policy. The network receives sensor values from the environment such as positional and velocity values of body parts, and outputs torque commands to the actuators.
The networks use tanh activation functions at each layer, and the  sizes are provided in \Cref{tab:params}.

The MAP-Elites algorithm uses a discretised behavioural descriptor space, divided into 1024 cells (centroids) using a CVT-based clustering method \citep{cvt}. 
We apply an iso+line mutation operator, with an isotropic noise standard deviation of $\sigma_{\text{iso}} = 0.005$ and directional line noise $\sigma_{\text{line}} = 0.05$ \citep{isoline}.
Parameters for SMOL-based schedules are mentioned in the Methodology section (1.5 to 1 for SMOL, 0.5 to 1 for SMOL-Reverse, 0.7 to 1.4 to 1 for SMOL-Human). These schedules are shown in \Cref{fig:schedules}.
All experiments are run for a total of $10^7$ evaluations with 7 to 10 random seeds each.
We report the statistical significance of our results using a one-sided Mann-Whitney U test \citep{mann1947test}. 
Additional task-specific hyperparameters, such as population size, descriptor types, and fitness functions, are set from usual values in the literature and summarised in \Cref{tab:params}.

\subsection{Tasks}

We benchmark our method on three Brax locomotion tasks: Halfcheetah, Walker2D, and Ant Omni, visualised in \Cref{fig:tasks}.
Halfcheetah and Walker2D are traditional forward locomotion tasks, where the goal is to maximise velocity over a fixed horizon of 1000 steps.
In these tasks, QD primarily functions as a way to discover stepping-stone behaviours that aid long-term optimisation.
Behavioural descriptors are based on foot contact patterns; that is, we define gait patterns as distinguished solely by their duty factors.

Ant Omni is a task designed to emphasize diversity.
The robot is encouraged to move in any direction from the origin, while minimizing energy usage as its fitness objective.
The behavioural descriptor is the agent’s final $(x, y)$ position after 100 environment steps.
Ant Omni, uses MAP-Elites to find controllers that reach all places surrounding the ant, while minimizing the energy spent in the fitness.
The interesting score of Ant Omni is hence the coverage, which captures the diversity of policies found by MAP-Elites.


\begin{table}[ht]
\centering
\caption{Hyperparameters for each task.}\label{tab:params}
\begin{adjustbox}{width=0.45\textwidth}
\begin{tabular}{l|ccc}
\toprule
\textbf{Hyperparameter} & \textbf{Halfcheetah} & \textbf{Walker2D} & \textbf{Ant Omni} \\
\midrule

Population size     & 8192            & 8192            & 1024            \\
Environment steps   & 1000           & 1000           & 100            \\
Descriptor space    & Foot contact & Foot contact & Final position \\
Fitness    & Velocity & Velocity & Energy used \\
Hidden layers        & [64, 64]       & [64, 64]       & [256, 256]       \\


\bottomrule
\end{tabular}
\end{adjustbox}
\end{table}

\subsection{Metrics}
During the QD optimisation run, we record results at every step using the following two primary metrics, computed on the current MAP-Elites archive using the current $\alpha$ value. We report the \textbf{Max Fitness} as the highest task performance achieved by any solution in the archive, and the \textbf{Coverage} as the  proportion of archive cells filled with a solution, reflecting the diversity of the archive. 
We do not report QD-score \cite{book_chapter} for conciseness, as results are similar to those reported with max fitness and coverage.


We want to highlight that as the $\alpha$ value changes the physics of the simulator, the fitness score and behaviour of policies may differ from their evaluation on the standard environment with $\alpha = 1$. For example, $\alpha > 1$ will give more power to the robot actuators, allowing it to move faster and hence reach higher fitness values in HalfCheetah or Walker2D. This is why we enforce $\alpha = 1$ for the last $10\%$ of the run of all schedules, noted as a vertical dashed line. Past the dashed line, all solutions are evaluated on the same environment, and metrics can be compared across methods. For metrics ahead of the dashed line, differences in performance may reflect difference in $\alpha$ as much as difference in learning performance.

\section{Main results}
\label{sec:main-results}


We first compare SMOL, the SMOL-Reverse and SMOL-Human variants to the standard schedule of Vanilla MAP-Elites. 
In the default schedule, the actuator scaling factor $\alpha = 1$ is held constant throughout training, following the standard procedure in MAP-Elites.
However, as in the other schedules, we still reevaluate the archive after each batch to account for evaluation noise and maintain methodological consistency.

\begin{figure*}
    \centering
    \includegraphics[width=0.9\linewidth]{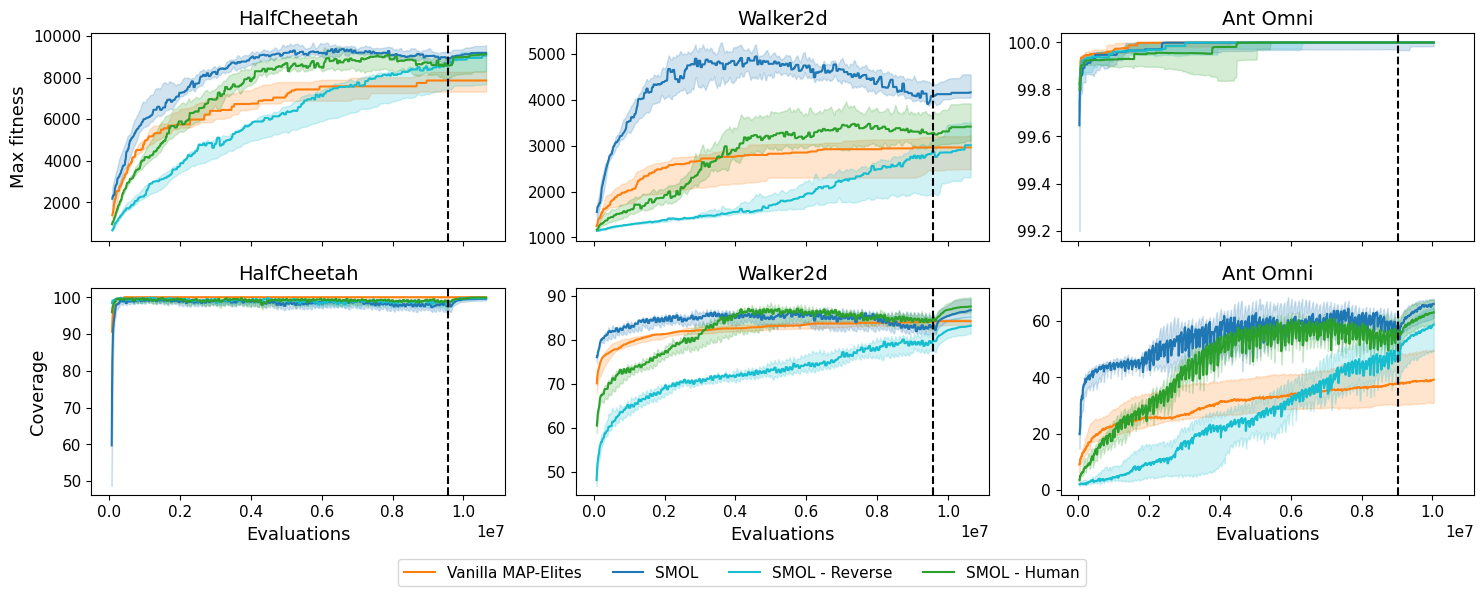}
    \caption{Performance of main experiments. Lines show the median values of max fitness and coverage. Shaded regions indicate the interquartile range. The vertical dashed line marks the point where the torque scaling factor $\alpha$ is fixed to 1 for all methods, ensuring a fair comparison in the final evaluation phase (see Metrics subsection).}
    \label{fig:main-results}
\end{figure*}

\begin{table}[h]
  \centering
    \input{tables/pvalues}
  \caption{Median final scores and P-values of U-tests, with the alternative hypothesis that the column method is better than the row method. P-values $<0.05$ in bold, meaning the column method statistically outperforms the row method.}
  \label{tab:pairwise_pvalues}
\end{table}

\Cref{fig:main-results} shows the median performance of each torque schedule across the three environments, with p-values of the Mann-Whitney U test provided in \Cref{tab:pairwise_pvalues}. 
The jagged lines in plots reflect drops in coverage and fitness caused by archive reevaluation: solutions which perform worse on the new physical constraints are removed.
We analyse this phenomenon in more detail in the \nameref{sec:ablations} section. 
We can also observe that fitness and coverage curves for SMOL tend to reach high values before going down: this is partly due to the increase in power, that makes the robot more powerful and faster. We hence use the last 10\% of the SMOL runs (after the vertical dashed line) and the statistical tests on final values for our analysis.

The two key takeaways from \Cref{fig:main-results} are: (1) the SMOL schedule outperforms other methods across all environments, and (2) all schedules match or exceed the performance of the default scaling (fixed-$\alpha$) MAP-Elites baseline.

In the Ant Omni task, all schedule-based methods reach a much higher coverage on the final archive than the baseline that used the same physics for the whole training run. 
Max fitness is similar across all baselines, as it is trivially maximised by minimizing energy through inaction, making it uninformative in this context.
Similarly, all schedule-based methods outperform the fixed baseline in HalfCheeetah and Walker2D, although the difference is not statistically significant for SMOL-Reverse on Walker2D. They also reach similar or higher coverage values than the baseline on these tasks. Overall, these results show that torque schedules in the training phase improve both performance and behavioural diversity compared to standard MAP-Elites. 

With both SMOL and SMOL-Reverse outperforming MAP-Elites while they have opposite developmental schedules, it seems that the constant $\alpha=1$ approach is set in-between two advantageous settings and benefits from none of them. 
We hypothesize that this is because early high torque in SMOL supports broad exploration and accelerates the discovery of diverse behaviours, while low torque reduces the reachable behaviour space and hence increases the selective pressure in SMOL-Reverse.

When compared to the Reverse variant that considers a growing power-to-weight ratio, the SMOL schedule inspired by muscle development consistently performs best, with p-values lower than 0.05 (0.034 on Ant Omni and 0.0006 on Walker2D) or close to it (0.057 on HalfCheetah). 

Interestingly, SMOL keeps a higher coverage than Vanilla MAP-Elites on Walker2D for most of the run, showing the additional power helps discover new locomotion gaits that seem to provide better stepping stones for the end of the run. 

On the other hand, SMOL-Reverse reaches the same fitness value in Halfcheetah and coverage in Ant Omni as Vanilla MAP-Elites, but before the vertical dashed line. This means that with a less powerful body, the schedule of SMOL-Reverse allowed to find similar (and then better) solutions than the standard method with a fully functional body. Such approach, while outperformed by SMOL, could hence allow to generate a set of policies that can perform well on degraded hardware such as used robots. 



Finally, while the SMOL-Human variant is less performant than the SMOL schedule, it also consistently outperforms MAP-Elites. This performance highlights how robotic learning can benefit from biologically inspired schedules, although more complex schedules like SMOL-Human may require some additional tuning to find a more efficient curriculum. 
In our setting, the robot did not undergo any body transformations except for the actuator torque, which may explain why the schedule inspired by the simpler muscle scaling model works better than the one that takes into account puberty and other isometrics changes in development.

\section{Ablations}
\label{sec:ablations}

\begin{figure*}
    \centering
    \includegraphics[width=0.9\linewidth]{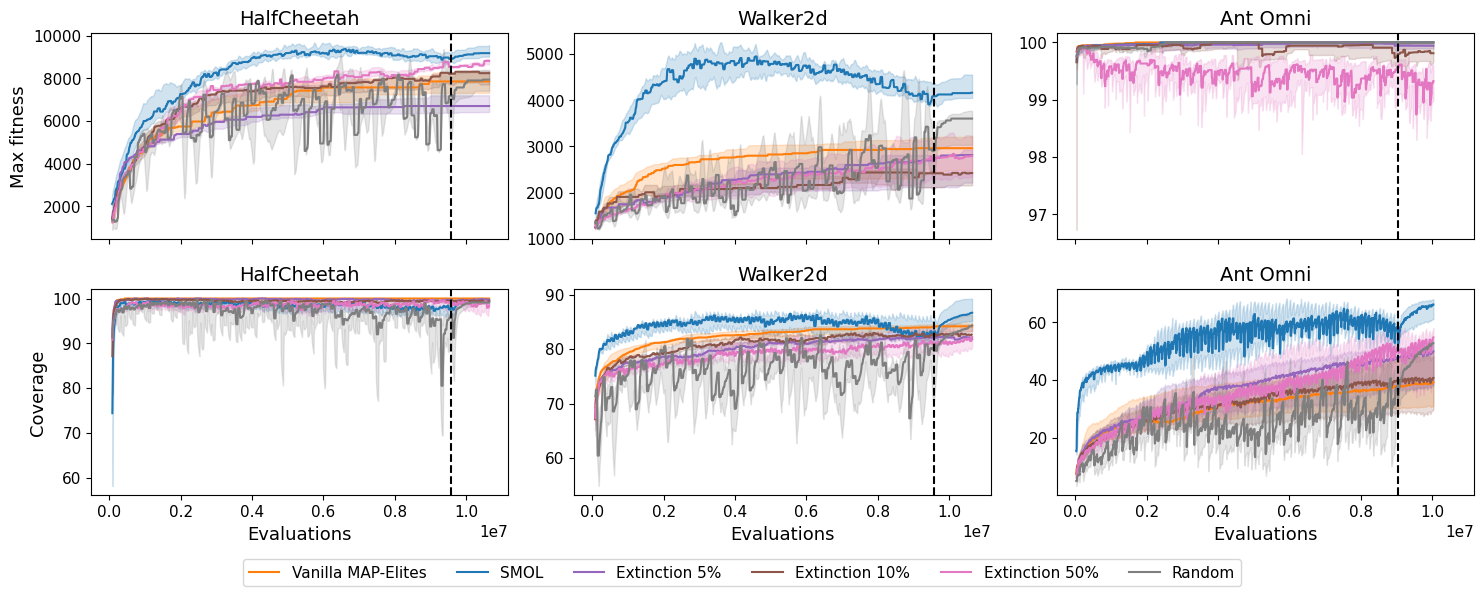}
    \caption{Performance of ablation experiments. Lines show the median values of max fitness and coverage. Shaded regions indicate the interquartile range. The vertical dashed line marks the point where the torque scaling factor $\alpha$ is fixed to 1 for all methods, ensuring a fair comparison in the final evaluation phase.}
    \label{fig:ablation-results}
\end{figure*}




Following the main results presented above, we conducted a series of ablation experiments to investigate alternative explanations for the observed performance gains of the SMOL baseline.
Specifically, we asked whether the benefits of our methods might arise from their impact on general training dynamics, such as domain randomisation or archive instability, rather than from the schedules themselves. 

\newpage
\subsection{Ablation Experiments}\


\paragraph{1) Random:} 
One hypothesis is that our schedules act as a form of domain randomisation, which is known to improve robustness in policy learning, particularly in sim-to-real transfer settings \citep{tobin2017domain, chen2021understanding}.
To evaluate this, we introduce a baseline where, at each schedule step (i.e., every $1\%$ of training), a new $\alpha$ is sampled uniformly at random from the range $[0.5, 1.5]$. 
As with the other schedules, we keep the final 10 phases where $\alpha = 1$ is fixed.
This setting introduces high variability in embodiment without a structured developmental trajectory.

\paragraph{2) Extinction:} 
Another possibility is that performance gains arise not from the schedule shape itself, but from the repeated reevaluation of the archive that occurs whenever $\alpha$ changes. 
These reevaluations can effectively act as soft extinction events, as they may remove or collapse behaviours that no longer function well under the new embodiment.
It has been observed that such disruptions can help MAP-Elites escape local optima and encourage continued exploration \citep{extinction, coiffard_overcoming_2025}.
Therefore, to isolate this effect, this baseline augments the default $\alpha = 1$ schedule with periodic extinction events.
Before each archive reevaluation, every solution in the archive is removed with probability $\sigma$, simulating a sudden loss of behavioural diversity.
By explicitly modelling extinction in a controlled setting without any changes to the robot’s body, we seek to test whether the benefits of our approach stem from biologically inspired power schedules, rather than from the destabilizing effects of archive turnover.
Following the observed drop of 5 to 10 points in coverage at every change in $\alpha$ when using SMOL, we test this ablation with three values of $\sigma$: $5\%$, $10\%$, and $50\%$.

\subsection{Ablation Results}
\Cref{fig:ablation-results} shows the performance of the ablation baselines across all three environments, compared with our method SMOL, with p-values of the Mann-Whitney U test provided in \Cref{tab:pairwise_pvalues} to compare final results.

Our first observation is that, across all tasks, the SMOL schedule continues to outperform the ablation baselines.
In Walker2D and Halfcheetah, SMOL achieves the highest max fitness and matches or exceeds all baselines on coverage.
In Ant Omni, SMOL again achieves better coverage than the ablations.
These results are both confirmed by the U-test results in \Cref{tab:pairwise_pvalues}, where SMOL statistically outperforms all baselines on all environments. In this table, SMOL-Human also shows good results, statistically outperforming most baselines in final scores on Ant Omni and Walker2D.
This reinforces the conclusion that SMOL’s effectiveness comes from the structured schedule itself, not incidental effects like simply changing the environment like domain randomisation or archive disruption by extinction. By starting with high torque, SMOL enables early exploration of a wide range of behaviours, which serve as strong stepping stones for later refinement—providing MAP-Elites with a developmental scaffold to build on as actuation becomes more limited.

Examining the Random baseline, we observe that performance is highly unstable in the early stages across all environments due to the constantly shifting physical constraints. As reported results are computed using the $\alpha$ value at this step, high scores before the vertical dashed line may be simply due to a more advantageous physics, hence the high interquantile range too. 
In Walker2D, this instability appears to aid exploration and provides robustness, resulting in improved max fitness over the fixed MAP-Elites baseline when they both train on the real environment at the end.
However, in Halfcheetah, this Random baseline yields similar fitness to the baseline and in Ant Omni, it results in the lowest coverage.
These results suggest that unstructured variations to physical parameters can occasionally help but may also be unreliable and detrimental to consistent progress. Randomly changing the environment leads to large shifts in the fitness and behaviour functions, which selects for strains of solutions that can adapt to any future environment but not specialize. With SMOL, solutions are more progressively selected to serve as stepping stones towards policies that perform well on the standard environment. 

Turning to the Extinction baselines, the results are less clear cut.
In Walker2D, all extinction rates produce similar results to MAP-Elites on both metrics.
In Halfcheetah and Ant Omni, the most aggressive extinction setting ($\sigma = 50\%$) leads to improved performance.
This supports previous findings that periodic archive disruption can encourage exploration \citep{extinction, coiffard_overcoming_2025}.
However, SMOL still clearly outperforms all extinction variants, indicating that extinction alone cannot explain its effectiveness.
Moreover, the drop in coverage seen in the main results (\Cref{fig:main-results}) after each archive reevaluation is roughly 5 to 10 percent points, suggesting that SMOL’s reevaluation effect is closest to the 5\% or 10\% extinction variants.
Yet SMOL significantly outperforms these baselines in all environments, providing further evidence that the schedule’s structure is the primary driver of improved learning and diversity.

\section{Conclusion}

In this work, we study how biological developmental changes can inspire new training schedules in evolutionary robotics. We focus on the difference in body size between young animals or humans and their adult counterparts, and posit that the ontological scaling of muscle power-to-weight ratio may play a part in the usefulness of early-life skill development for adult locomotion. Focusing on robotic legged locomotion, we develop new schedules of actuator torques inspired by muscle development and human growth, that prove to outperform training directly with the final model. 

While this study focuses on one specific change in the body through the force output, we argue that QD methods are particularly adapted for developmental learning, as humans and animals tend to find many diverse ways to solve the same problem during their lifetime, which informs them when they get older. While QD methods have been used to adapt to sudden damage such as a broken leg \citep{nature}, methods like SMOL could be used for progressive damage adaptation, which would be better adapted for ageing robots or slow degradation. 

A key difference between QD approaches and the way animals and humans learn is in how we focus on useful skills like walking fast, while MAP-Elites will spend part of its evaluation budget on hopping faster on each foot to cover the archive. In future work, we will examine how methods like Quality with Just Enough Diversity (JEDi) \citep{jedi}, which estimates which behaviour will lead to higher scores, would behave when integrated with SMOL schedules.

Finally, while our findings concern robotics, they invite a parallel with developmental biology: the benefits of a changing power-to-weight ratio for learning locomotion in robots may hint at a similar influence on the evolutionary value of play, offering young animals opportunities to refine these skills early in life. We hope that future research in biology could provide further insights to guide robotics in understanding how developmental changes impact learning.


\section{Acknowledgements}
This work was supported by the DARPA and its Learning Introspective Control (LINC) program.

\footnotesize
\bibliographystyle{apalike}
\bibliography{reference}

\end{document}

%% file: tables/pvalues.tex
\begin{subtable}{0.45\textwidth}
  \centering
  \begin{adjustbox}{width=\textwidth}
    \begin{tabular}{l|c|c|c|c|}
    \toprule
     & Median & \multicolumn{3}{c|}{p-values}  \\
     & score & SMOL & SMOL - Reverse & SMOL - Human \\
    \midrule
    SMOL & \textbf{66} & - & 9.714e-01 & 6.377e-01 \\
    SMOL - Reverse & 58 & \textbf{3.352e-02} & - & 8.673e-02 \\
    SMOL - Human & 63 & 3.891e-01 & 9.246e-01 & - \\
    Vanilla MAP-Elites & 39 & \textbf{5.971e-04} & \textbf{2.258e-02} & \textbf{1.101e-03} \\
    Random & 53 & \textbf{2.705e-03} & 1.061e-01 & \textbf{7.010e-03} \\
    Extinction 5\% & 50 & \textbf{1.874e-04} & \textbf{4.443e-02} & \textbf{1.799e-03} \\
    Extinction 10\% & 40 & \textbf{8.249e-05} & \textbf{8.629e-03} & \textbf{2.914e-04} \\
    Extinction 50\% & 55 & \textbf{3.353e-03} & 2.137e-01 & \textbf{1.287e-02} \\
    \bottomrule
    \end{tabular}
  \end{adjustbox}
  \caption{Ant Omni (coverage)}
\end{subtable}

\begin{subtable}{0.45\textwidth}
  \centering
  \begin{adjustbox}{width=\textwidth}
    \begin{tabular}{l|c|c|c|c|}
    \toprule
     & Median & \multicolumn{3}{c|}{p-values}  \\
     & score & SMOL & SMOL - Reverse & SMOL - Human \\
    \midrule
    SMOL & \textbf{9202} & - & 9.529e-01 & 9.604e-01 \\
    SMOL - Reverse & 9076 & 5.567e-02 & - & 4.299e-01 \\
    SMOL - Human & 9123 & \textbf{4.708e-02} & 6.045e-01 & - \\
    Vanilla MAP-Elites & 7857 & \textbf{1.436e-04} & \textbf{2.412e-02} & \textbf{1.670e-02} \\
    Random & 7943 & \textbf{1.332e-03} & 7.343e-02 & 9.491e-02 \\
    Extinction 5\% & 6704 & \textbf{3.497e-03} & 7.273e-02 & \textbf{1.818e-02} \\
    Extinction 10\% & 8246 & \textbf{9.990e-04} & 5.315e-02 & 7.413e-02 \\
    Extinction 50\% & 8820 & \textbf{3.846e-02} & 2.409e-01 & 4.318e-01 \\
    \bottomrule
    \end{tabular}
  \end{adjustbox}
  \caption{HalfCheetah (Max fitness)}
\end{subtable}

\begin{subtable}{0.45\textwidth}
  \centering
  \begin{adjustbox}{width=\textwidth}
    \begin{tabular}{l|c|c|c|c|}
    \toprule
     & Median & \multicolumn{3}{c|}{p-values}  \\
     & score & SMOL & SMOL - Reverse & SMOL - Human \\
    \midrule
    SMOL & \textbf{4116} & - & 9.995e-01 & 9.925e-01 \\
    SMOL - Reverse & 3099 & \textbf{6.165e-04} & - & 8.395e-02 \\
    SMOL - Human & 3610 & \textbf{8.906e-03} & 9.257e-01 & - \\
    Vanilla MAP-Elites & 3103 & \textbf{2.774e-05} & 3.468e-01 & \textbf{7.558e-03} \\
    Random & 3602 & \textbf{2.620e-03} & 8.125e-01 & 3.295e-01 \\
    Extinction 5\% & 2812 & \textbf{1.616e-04} & 1.886e-01 & \textbf{9.615e-03} \\
    Extinction 10\% & 2422 & \textbf{2.424e-02} & 2.917e-01 & \textbf{3.434e-02} \\
    Extinction 50\% & 2813 & \textbf{5.495e-04} & 2.447e-01 & \textbf{3.883e-02} \\
    \bottomrule
    \end{tabular}
  \end{adjustbox}
  \caption{Walker2d (Max fitness)}
\end{subtable}